# Fast and reliable stereopsis measurement at multiple distances with iPad

Manuel Rodriguez-Vallejo[1*], Clara Llorens-Quintana[3], Diego Montagud[3], Walter D. Furlan[3] and Juan A. Monsoriu[2]

[1] *Qvision, Unidad de Oftalmología Vithas Hospital Virgen del Mar, 04120, Almería*

[2] *Centro de Tecnologías Físicas, Universitat Politècnica de València, 46022 Valencia, Spain*

[3] *Departamento de Óptica, Universitat de València, 46100 Burjassot, Spain*
[*]*Corresponding author: manuelrodriguezid@qvision.es*

1. ABSTRACT

**Purpose:** To present a new fast and reliable application for iPad (ST) for screening stereopsis at multiple distances.

**Methods:** A new iPad application (app) based on a random dot stereogram was designed for screening stereopsis at multiple distances. Sixty-five subjects with no ocular diseases and wearing their habitual correction were tested at two different distances: 3 m and at 0.4 m. Results were compared with other commercial tests: TNO (at near) and Howard Dolman (at distance) Subjects were cited one week later in order to repeat the same procedures for assessing reproducibility of the tests.

**Results:** Stereopsis at near was better with ST (40 arcsec) than with TNO (60 arcsec), but not significantly ($p = 0.36$). The agreement was good ($k = 0.604$) and the reproducibility was better with ST ($k = 0.801$) than with TNO ($k = 0.715$), in fact median difference between days was significant only with TNO ($p = 0.02$). On the other hand, poor agreement was obtained between HD and ST at far distance ($k=0.04$), obtaining significant differences in medians ($p = 0.001$) and poorer reliability with HD ($k = 0.374$) than with ST ($k = 0.502$).

**Conclusions:** Screening stereopsis at near with a new iPad app demonstrated to be a fast and realiable. Results were in a good agreement with conventional tests as TNO, but it could not be compared at far vision with HD due to the limited resolution of the iPad.

## 2. INTRODUCTION

Stereopsis is a measure of the visual perception of three-dimensional space. It is based on the binocular retinal disparity and it is included in vision examination of adults and children (American Optometric Association 1994a; American Optometric Association 1994b). The smallest binocular disparity that can be detected is known as stereoacuity or stereothreshold and is recorded in seconds of arc (arcsec) (Westheimer 2013). The Howard–Dolman (HD) two-rod apparatus is considered the gold standard to measure stereoacuity; however, this device has important drawbacks to be used in clinical practice: it is time-consuming, it should be performed at distance to avoid monocular clues, and requires that the subject makes a complex motor task with repeated measurements (Saladin 2005). Three types of tests are preferred in clinical practice depending on the type of stereopsis in which they are based: local stereopsis with contours or bars, global stereopsis with random dots, and a combination of both known as real stereopsis (Fricke et al. 1997). Even though all these tests are appropriated for vision screening in clinical practice, results may differ between them since they have not been designed to measure continuous stereoacuity as HD (Simons 1981). On the other hand, they are mainly used to detect, by discrete step sizes, whether there is any abnormality that affects binocular vision.

The most popular stereopsis tests are conducted at near. However, in the last decade distance stereotesting has been also suggested as a good screening procedure, highly sensitive to small refractive error changes, heterophorias and strabismus (Wang et al. 2010), and Snellen visual acuities under 20/25 (Rutstein and Corliss 2000). Even though distance stereopsis can provide additional information not detectable by near tests, the fact is that there are a lot of clinicians who have not incorporated this testing in their routines, perhaps because there are few tests designed to measure distance stereopsis, which are also very expensive.

New vision tests have been developed with the emerge of portable screens such as IPad or Android Tablets and smartphones (Zhang ZT, Zhang SC, Huang XG 2013; Perera et al. 2015). Several of the advantages of computerized tests have been implemented in these portable devices: randomized letters, automated scoring, wide range of optotypes, bright screen calibration, normal population databases, remote connection, etc. (Rodríguez-Vallejo et al. 2015). Other applications to measure stereopsis for portable devices such as iPod have been proposed but for only using at one distance, in long time (3 minutes), and not compared with conventional stereotests (Hess et al. 2016). Thus, the introduction of versatile, fast and portable stereopsis tests which can be used at different distances is of primary importance. This is the main goal of this work, a new iPad application for stereopsis measurement is presented and its reliability is assessed against the results obtained with the HD and TNO tests.

## 3. MATERIALS (OR SUBJECTS) AND METHODS

In this section we first introduce the principles in which the new test was inspired. Thus, we begin with a brief review of the Howard Dolman and TNO test in order to put our analysis in a proper framework.

*Howard Dolman*

HD principle is schematized in Fig. 1A. Two vertical rods are seen in front of an empty field, one of them (O) is fixed and the other one (O') is movable back and forth along a lane. The rods are seen by the observer at 3 meters and the task is to align the movable rod, with a string attached to it, until the observer perceives that both rods are at the same distance. The stereoacuity ($\gamma$ in radians is obtained in a continuous scale from the measurement of the relative distance between the two rods along the line of sight ($\Delta z$):

$$\gamma = \frac{a\Delta z}{z^2} \qquad (1)$$

where z is the distance from the observer to the fixed rod and *a* is the interpupillary distance (see Fig. 1A) (Westheimer 2013).

The HD (Bernell Corporation)(Bernell 2014) used in this study has a continuous scale up to 73 arcsec, but measurements above 66 arcsec were considered outside of the instrument limits (OL), or suspended stereopsis.

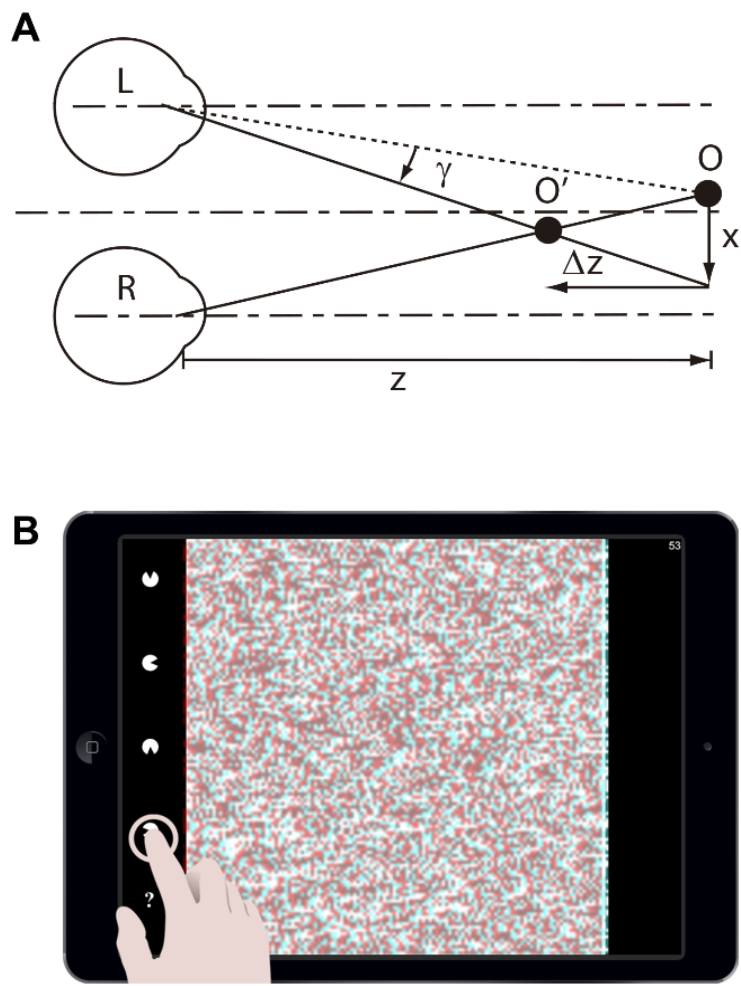

*Figure 1. (A) Two objects scheme for computing the stereoacuity. (B) Screen capture of the app during the measurement process.*

*TNO*

The TNO is a random dot test to measure global stereopsis at near (40 cm) with anaglyph eyeglasses (van Doorn et al. 2014). It consists of seven plates: three for screening purposes, a suppression test, and three plates to measure stereoacuity. In this last case, the task is to identify the position of a missing section of a circle that appears at one of four possible orientations. These figures are presented at six different depth levels corresponding to a disparity in the range from 15 to 480 arcsec (see Table 1) (van Doorn et al. 2014).

*Table 1. Discrete steps of measurement with ST and TNO at near. A variable conversion from arcsec to an ordinal scale of five levels was performed for computing the Cohen's k with quadratic weights.*

| Level (Range) | ST (arcsec) | TNO (arcsec) |
|---|---|---|
| 5 (OL*) | OL* | OL* |
| 4 (278-480) | 397 | 480 |
|  | 357 |  |
|  | 318 |  |
|  | 278 |  |
| 3 (159-240) | 238 | 240 |
|  | 199 |  |
|  | 159 |  |
| 2 (79-120) | 119 | 120 |
|  | 79 |  |
| 1 (15-60) | 40 | 60 |
|  |  | 30 |
|  |  | 15 |

* Outside Device Limits (OL). The subject cannot resolve the stimulus with the TNO or ST.

*iPad-Stereo Test*

The iPad Stereotest (ST) we propose was developed with pure ActionScript 3.0 programming language for mobile devices and then compiled for IOS with Adobe Flash Builder (Adobe Systems, Inc.). Two identical arrays of random colored dots (one in red and one in cyan) are displayed in such a way that each array is visible with one of the patient's eyes when it wears anaglyph eyeglasses (red filter on the left eye). Some ordered dots inside a circle with a gap (similar to those in plates V, VI, and VII in the TNO test), are laterally displaced to produce fixed amounts of binocular disparity degree in arcsec (see the values in Tables 1 and 2). The binocular fusion of both patterns simulates a stereoscopic object when the disparity is crossed. As can be seen in Fig. 1A, if the displacement between the corresponding dots images at the reference (iPad) plane, is $x$; the stereoacuity can be expressed as $\gamma \approx x/z$ provided that in practice $z \gg \Delta z$.

Considering that, due to screen's pixels density (*SPD;* in pixels per inch (ppi)), the lateral displacement (*x*) is limited by pixel size, the stereopsis, can be computed (in radians) in terms of the *SPD* as:

$$\gamma = \frac{i}{SPD \; z_0} \qquad (2)$$

where *i* is the number of pixels of displacement corresponding to the distance $z_0$. In order to evaluate the same level of stereopsis at multiple distances $z_j$ greater than $z_0$, an integer multiplicative constant must be inserted on the right member of Eq. (2) such that $k = z_j/z_0$. However, the display resolution imposes a limit on the finest value that can be measured. For the iPad *retina* this limit is 40 arcsec for a presentation distance of 0.5 m. The *SPD* value is automatically recovered from the tablet by means of the programming code in order to avoid the need to calibrate the stimulus size with an external rule. On the other hand, the size of the random dots is variable with the presentation distance in a way that each dot subtends an angle of 1.32' at all distances which corresponds to a minimum visual acuity of 0.125 logMAR. The stereoscopic stimulus size is constant and subtends 1.88º at 3 m. Fig. 1B shows a screen capture of the app during the trial.

The stereoacuity scale is divided in ten discrete steps, being the lower value of the scale limited the pixel size and the following values of the scale are obtained by increasing one pixel of disparity between the images. An automated method to achieve the threshold was included in the app. In it, the level of stereopsis goes one level down with each right answer until the subject fails, then stereopsis goes one level up after the patient fails again. Stereo-threshold was considered the last level on which subject's response is correct after the first fail. The time spent for complete the trial is around 30 seconds.

*Subjects and Procedures*

Sixty-five subjects (mean age: 27.7 ± 7.2 years) were requited during a vision screening in the University of Valencia. Informed consent was obtained for each subject and the research was conducted in accordance with the principles laid down in the Declaration of Helsinki. Previous to stereopsis measurements, monocular visual acuity and cover test were evaluated; as well as objective refraction and interpupilar distance with WAM-5500 (Grand Seiko Co., Ltd., Hiroshima, Japan) (Sheppard and Davies 2010). Exclusion criteria were ocular diseases, strabismus, monocular visual acuity under 0.1 logMAR, a difference of 0.1 logMAR between both eyes with best compensation, and a residual spherical equivalent higher than ±0.50D from the objective value measured with the WAM-5500 with the subject wearing the habitual correction in spectacles or contact lenses.

All measurements were undertaken in the same room under artificial lighting conditions: 285 lux (LX1330B luxmeter). The device used to perform this research was an iPad third generation with retina display (2048-by-1536-pixel resolution and 264 ppi) with brightness at 100%; which corresponded to 342 cd/m² for white color (Spyder4Elite colorimeter).

Stereopsis was first measured at 3 m with the HD, each value was obtained with a psychophysical method, averaging the absolute values of six measures, three of them obtained starting with the movable rod ahead of the fixed rod (descending) and other three starting with the movable rod behind of the fixed rod (descending) (Ehrenstein and Ehrenstein 1999). Then, stereoacuity was measured with ST at the same distance with ten different threshold levels of stereoacuity (see Table 2).

Table 2. Discrete steps of measurement with ST and reorganization of HD measurements from a continuous scale to a range for computing the agreement with the Cohen's k with linear weights.

| Level (Range) | ST (arcsec) | HD (arcsec) |
| --- | --- | --- |
| 11 (OL*) | OL* | OL* |
| 10 (66-63) | 66 | 66-63 |
| 9 (56-62) | 60 | 56-62 |
| 8 (55-50) | 53 | 55-50 |
| 7 (49-44) | 46 | 49-44 |
| 6 (43-37) | 40 | 43-37 |
| 5 (36-30) | 33 | 36-30 |
| 4 (29-24) | 26 | 29-24 |
| 3 (23-18) | 20 | 23-18 |
| 2 (10-17) | 13 | 10-17 |
| 1 (9 – 0) | 7 | 9 - 0 |

\* Outside Device Limits (OL). The subject cannot resolve stimulus with the ST or the movable bar is above 66 arcsec with the HD.

Once distance stereopsis was evaluated, the patient was positioned at 50 cm from the iPad and the near stereoacuity was measured for other ten different threshold levels (see Table 1).

Finally, the procedure was completed by testing each subject with the TNO at 40 cm under warm light of 945 lux. Subjects were cited a week after the first session to repeat all the procedures described above in order to assess the reproducibility of each test.

*Statistical Analysis*

Non-parametric statistics were used because of the non-normal distributions of the variables. Median significant differences between instruments at first day and between days for the same instrument were evaluated with the Wilcoxon Signed Rank test whereas the agreement and reproducibility was computed with the Cohen's k with linear weights for distance stereopsis and quadratic weights for near stereopsis, the reason for linear weights at distance and quadratic at near was because stereoacuity steps are increased in an approximated linear way at distance but not at near. To evaluate the agreement at far, taking into account that HD measures stereopsis in a continuous scale whereas ST uses discrete step sizes, HD data were discretized in a set of values closer to the nearest stereoacuity in ST scale (see Table 2). On the other hand, for measurements of the stereoacuity at near a reorganization of data was performed in order to compare results from TNO and ST, since ST and TNO use different discrete steps. In this case, results were recoded to an ordinal scale from 1 to 5 depending on the stereopsis achieved with TNO and ST (see Table 1). Statistical analyses were performed using the SPSS software (ver. 20; SPSS Inc., Chicago, IL, USA) and MedCalc (ver. 12.7; MedCalc Inc., Belgium). The significance was accepted at the p<.05 level.

## 4. RESULTS

*Near Stereopsis*

The results for near stereopsis are shown in Fig. 2. Median stereopsis was slightly better for ST (40 arcsec) than for TNO (60 arcsec) even though no statistically significant differences were found in the comparison of medians between both tests ($p = 0.36$). A total of 84.6% of subjects achieved the finest level of 40 arcsec with the ST whereas 63.1% perceived up to 60 arcsec value with TNO. From the latter group, only six subjects perceived the 30 arcsec plate and one subject the 15 arcsec plate. The cumulated percentage of subjects who achieved the second level of stereopsis was closer for both tests, 83.1% with TNO and 92.3% with ST, and were equal at third level (see Fig 2A).

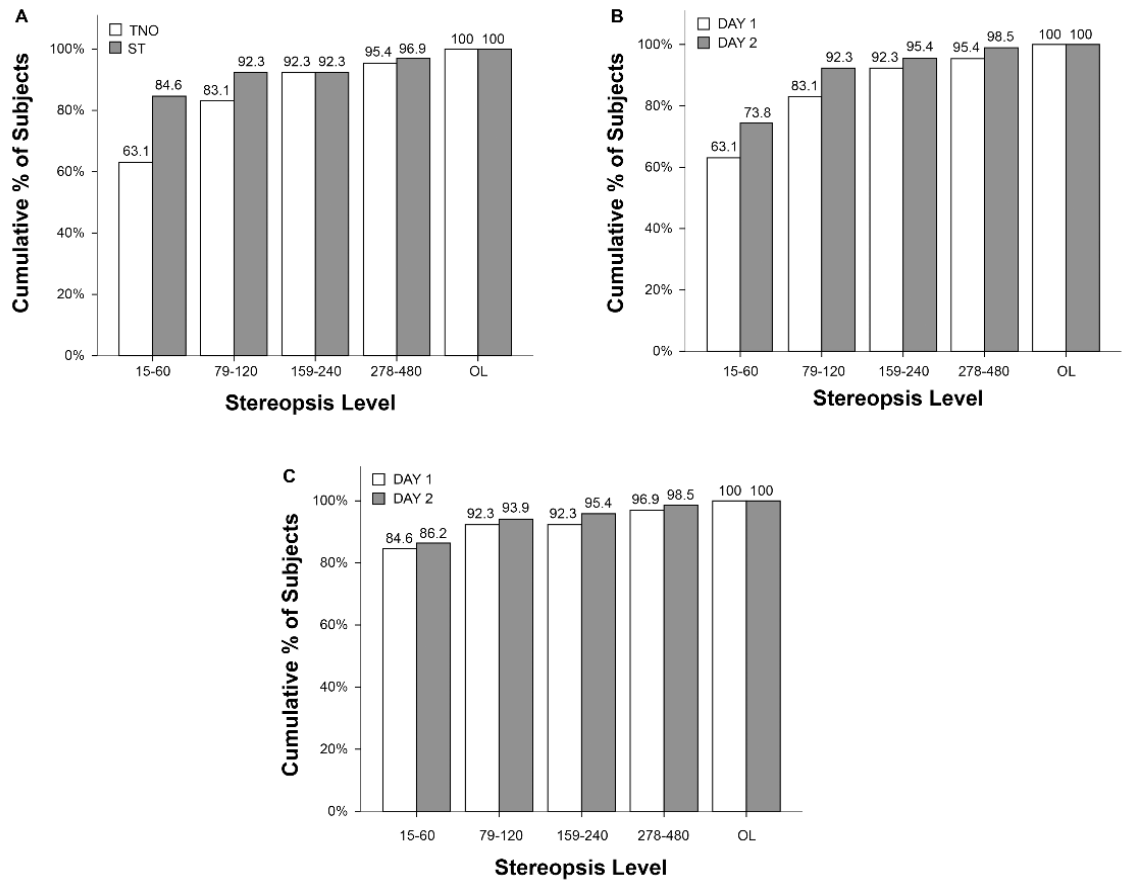

*Figure 2. Cumulative percentage of subjects who achieved a value of stereopsis at near inside the range of each level for (A) TNO and ST at first day, (B) TNO at both days, and (C) ST at both days.*

The Cohen's k for quadratic weights resulted in substantial agreement k=0.604 95% CI (0.300, 0.908) between both instruments according with Landis & Koch criteria (Landis and Koch 1977). Only one subject failed with ST and TNO at both days whereas two subjects failed the TNO at first day but not the ST. On the contrary way, one subject failed the ST but not the TNO also at first day. All subjects except the one mentioned above passed both near stereopsis tests at the second day (see Table 4).

Statistically significant differences were found for the median of both days with the TNO (p= 0.02) but not with the ST (p= 0.301) (see Table 3). In addition reproducibility was better with ST (k=0.801, IC95% [0.584, -1.000]) than with TNO (k=0.715, IC95% [0.520, -0.909]). This poorer reproducibility of TNO was more remarkable for the first two levels of stereopsis (see Figs. 2B and 2C)

*Far Stereopsis*

The results for far distance stereopsis are shown in Fig.3. A Wilcoxon signed rank test revealed statically significant differences between medians of the measurements obtained with HD and ST (p=0.001). Fig. 3A shows that 82% of subjects achieved stereopsis between 18 and 23 arcsec; however, this percentage of subjects was not reached up to the range of 56-62 arcsec with ST. Therefore, ST underestimates the stereoacuity with regard to the HD as can be seen in medians at Table 3. The Cohen's k with linear weight was run to determine if there was agreement between stereoacuity obtained with HD and ST. Slight agreement was found between both instruments according with Landis & Koch criteria (Landis and Koch 1977): k=0.040 95% CI [-0.063, 0.142].

All patients aligned the bars of the HD inside of instrument limits, however seven subjects failed with ST in both days (see Table 4). With regard reproducibility, no significant differences in median were found between days for both tests (see Table 3) even though better reproducibility was obtained with ST (k= 0.502 95%CI [0.356-0.648]) than with HD (k= 0.374 95%CI [0.185-0.564]). (see Figs. 3B and 3C)

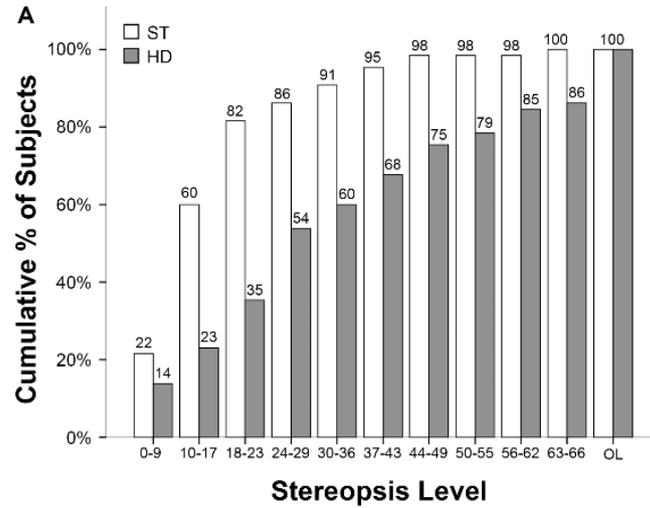
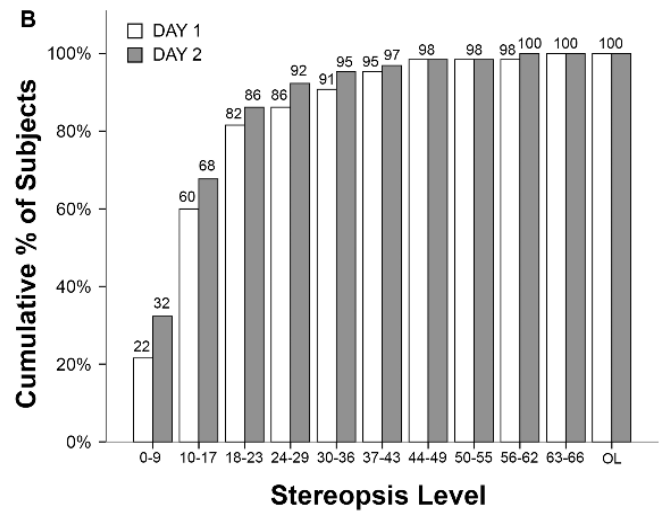
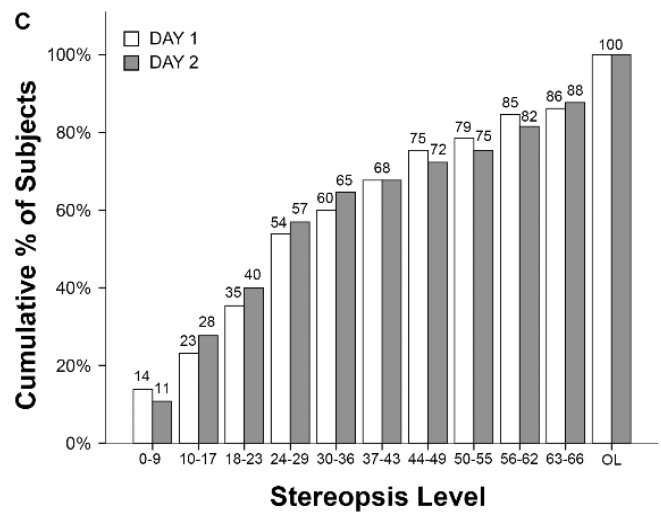

*Figure 3. Cumulative percentage of subjects who achieved a value of stereopsis at distance inside the range of each level for (A) HD and ST at first day, (B) HD at both days, and (C) ST at both days.*

*Table 3. Reproducibility analysis between days with distance and near stereo-tests. Wilcoxon and Cohen's K were computed in order to assess the difference in medians and the concordance, respectively.*

|  |  | Day | Median (arcsec) [interquartile range] | Wilcoxon | Cohen's k [95% CI] |
|---|---|---|---|---|---|
| Distance | HD | 1 | 16 [10 to 22] | z=-1.670, p= .095 | 0.374 [0.185-0.564] |
|  |  | 2 | 12 [8 to 20] |  |  |
|  | ST | 1 | 26 [20 to 46] | z=-.992, p= .321 | 0.502 [0.356 -0.648] |
|  |  | 2 | 26 [13 to 53] |  |  |
| Near | TNO | 1 | 60 [60 to 120] | z=-3.112, p= .02 | 0.715 [0.520-0.909] |
|  |  | 2 | 60 [60 to 90] |  |  |
|  | ST | 1 | 40 [40 to 40] | z=-1.034 p= .301 | 0.801 [0.584-1.000] |
|  |  | 2 | 40 [40 to 40] |  |  |

*Table 4. Results with each test for subjects who failed or whose stereopsis was outside the device limits (OL) with at least one of the tests.*

|  | Distance | | | | Near | | | |
|---|---|---|---|---|---|---|---|---|
|  | HD (arcsec) | | ST (arcsec) | | TNO (arcsec) | | ST (arcsec) | |
| Subject id | D1 | D2 | D1 | D2 | D1 | D2 | D1 | D2 |
| 17 | 7 | 7 | OL | 13 | 120 | 120 | 40 | 40 |
| 26 | 9 | 4 | 53 | 26 | OL | 480 | 40 | 79 |
| 27 | 23 | 13 | OL | OL | 60 | 60 | 40 | 40 |
| 29 | 15 | 11 | OL | OL | OL | OL | OL | OL |
| 31 | 28 | 8 | 60 | 40 | 240 | 120 | OL | 397 |
| 32 | 13 | 20 | 40 | OL | 240 | 120 | 397 | 159 |
| 35 | 34 | 9 | OL | OL | OL | 240 | 357 | 397 |
| 44 | 2 | 3 | OL | OL | 120 | 60 | 119 | 40 |
| 48 | 8 | 8 | OL | OL | 60 | 79 | 30 | 40 |
| 52 | 25 | 10 | OL | 46 | 60 | 60 | 40 | 40 |
| 62 | 22 | 12 | OL | OL | 60 | 60 | 79 | 40 |
| 64 | 39 | 43 | OL | OL | 60 | 120 | 40 | 40 |

## 5. DISCUSSION

The goals of this study were to validate a new stereopsis test for iPad and to assess its reproducibility in comparison with TNO at near and HD at distance. The stereopsis at near is widely used in clinical practice in order to improve paediatric vision screening programs, to provide an overall indication of binocularity, and to monitor binocularity after vision therapy or after monovision (Fricke et al. 1997; Rutstein et al. 2015). On the other hand, the measurement of distance stereopsis is uncommon in clinical practice even though it has been suggested that may be more effective than near stereopsis testing in screening for binocular vision disorders, reduced visual acuity, and uncorrected refractive errors (Rutstein and Corliss 2000). Particularly, distance stereopsis is useful in cases of intermittent exotropia on which patients pass near stereoacuity but fail at distance (Holmes et al. 2009), therefore it has been used for assessing the effect of surgery on binocular restoration in adolescents with this condition (Feng et al. 2015). In our opinion, the reasons why distance stereopsis is not widely used in clinical practice are that there are not many commercial tests to measure distance stereopsis, they are considerably expensive in comparison with near stereotests and two different tests are required to measure stereopsis at near and distance. Therefore, the use of a single and non-expensive test for measuring stereopsis at multiple distances is of major value.

TNO is one of the most popular tests in clinical practice and it has been used to determine how stereoacuity decreases with age (Lee and Koo 2005), in preschool screenings (Friendly 1978), and it has been compared with other tests (Simons 1981). Since ST and TNO do not measure the stereoacuity with the same discrete steps, we applied a transformation of variables by dividing subjects in groups depending on stereoacuity ranges. The agreement between both tests was good which means that ST and TNO can be used interchangeably for screening purposes; moreover, reproducibility of ST was better than TNO. One reason for this difference could be the illumination of the screen: the retro-illuminated screen of the iPad could make easier the fusion of images and the perception of the stereoscopic image than the TNO, even though it was very well illuminated in our experiment. This may be also the reason for the higher cumulative percentage of subjects for the ST up to the 79-120 arcsec range. The evidence of this study points towards the idea that ST could be more useful during screening programs on which room lighting is under 945 lux, conversely in exteriors under extreme sunlight conditions iPad screen brightness would not be enough to perform the test properly. Moreover, the measurement of stereopsis at very high luminances would not be adequate because stereoacuity may suffer from a decrement of performance in this condition (Westheimer 2013). Particular attention should be paid to avoid reflections from any overhead glare sources when we use the iPad in opposite to the TNO on which overhead light could be required (Black et al. 2013).

One of the strengths of ST is the possibility to measure stereopsis at multiple distances maintaining the same sensitivity of stereoacuity. This makes the app especially useful to measure stereopsis at near, intermediate and distance vision with multifocal contact lenses or after cataract surgery with multifocal refractive intraocular lenses (Rutstein et al. 2015), even though an understimation with diffractive intraocular lenses might be found due to be a wavelength-based stereotest (Varón et al. 2014).

Differences between results at near and distance might be attributable to the variation in the angular size of the stimulus between both distances. This could be of great importance in the evaluation of micro-strabismus in children because it has been demonstrated that stimulus size matters and central areas of suppression may difficult the perception of small stimulus (Pageau et al. 2015). As the angular size of the stimulus decreases when the iPad is moved away from the observer, a patient with central suppression might fail the test at distance but not at near on which the stimulus subtends a high angle. Future versions will include the possibility to vary angular size of the stimulus dynamically to assess the extension of suppression.

The ST has been designed with a minimum distance between random dots corresponding at all distances to a visual acuity of 0.125 logMAR. Therefore, subjects who pass the test at distance should have a monocular visual acuity better than 0.1 logMAR. This would improve the speed of vision screenings by performing the ST test at distance and assuming the absence of visual acuities poorer than 0.1 logMAR without conducting a visual acuity test. Myopic refractive errors could be easily detected with ST at distance even though hyperopia may go undetected because the subject might accommodate like with a visual acuity test (Suryakumar and Allison 2015). Hess et al., with a similar random dot stereotest, also reported a loss in stereopsis with the reduction in visual acuity (Hess et al. 2016). For this reason for the validation of ST, we decided to include subjects with monocular visual acuity better than 0.1 logMAR in order to ensure that they would perceive properly the ST at distance.

The agreement of ST with HD at distance was low even though the reproducibility was better for ST. Gantz & Bedell assumed that the disagreement between thresholds using local (HD) and global stereotargets (like ST) can be explained by differences in the properties of the targets or by differences in the neural mechanisms that underlie the processing of local and global stereograms (Gantz and Bedell 2011). A person performing well on a global test will perform acceptably well on a local stereopsis test, but the reverse is not true (Saladin 2005). In local stereopsis, this is due in part to the presence of contours that provide assistance to the fusion and that there is not a need for an accurate motor control; conversely in global stereopsis it is required an accurate bifoveal fixation (Fricke et al. 1997). Although we expected to have not very high agreement with HD before the study we decided to use the HD for comparison because it is considered the gold standard for stereopsis measurement (Saladin 2005).

## 6. CONCLUSION

We have validated a new application to measure stereopsis with iPad which has the primary advantage of measuring stereopsis at multiple distances, being more reproducible than other current clinical stereotests such as TNO or HD, and spending less time than other apps (Hess et al. 2016). The main limitation of this new test is that stereoacuity levels depends on pixel size; therefore, the limit for the iPad Retina (264 ppi) at 0.5m is 40 arcsec. In addition, the stereoacuity steps change depending on the tablet on which the ST is presented achieving finest values of stereopsis at near with tablets or phones with highest SPD (i.e., finest stereopsis value would be 32 arcsec at near with iPhone 6 or iPad mini which have 326 ppi). If we want to measure properly stereopsis at distance it is important to note that some conditions such as small refractive errors, anisometropia or visual acuities poorer than 0.1 logMAR should be controlled. The test could be difficult to perceive at distance in comparison with the same at near because of the smaller background pattern or stimulus size. Future studies are needed with subjects with different ocular anomalies in order to determine the sensitivity and specificity of this new test in some of the applications that we have commented in the discussion.

## 7. ACKNOWLEDGEMENTS

This study was supported by the Ministerio de Economía y Competitividad and FEDER (Grant DPI2015-71256-R), and by the Generalitat Valenciana (Grant PROMETEOII-2014-072), Spain.

## 8. CONFLICT OF INTEREST

Rodríguez-Vallejo, M. has designed and programmed the stereopsis app which he currently distributes by the Apple Store with his own developer account. The other authors report no conflicts of interest and have no proprietary interest in any of the materials mentioned in this article.